\documentclass{article}
\usepackage[utf8]{inputenc}
\usepackage{fullpage}

\usepackage{biblatex}
\usepackage{graphicx}
\usepackage{xcolor}
\usepackage{amsmath}
\usepackage{amsfonts}
\usepackage{amssymb}
\usepackage{mathrsfs}
\usepackage{dsfont}
\usepackage{algorithm}
\usepackage{algpseudocode}
\usepackage{soul}
\usepackage[english]{babel}

\def\E{\mathbb{E}}
\def\R{\mathbb{R}}
\def\N{\mathcal{N}}

\def\D{\mathcal{D}}
\def\P{\mathbb{P}}

\def\GP{\mathscr{GP}}

\def\n1{^{n+1}}

\def\GP{\text{GP}}

\def\KG{\text{KG}}
\def\KGOS{\text{KG}_{\text{OS}}}
\def\KGMC{\text{KG}_{\text{MC}}}
\def\Opt{\text{{\tt Optimizer}}}

\def\GP{\mathcal{GP}}

\def\KG{\text{KG}}
\def\yn1{y^{n+1}}
\def\mun1{\mu^{n+1}}
\def\n1{^{n+1}}

\DeclareMathOperator*{\argmax}{arg\,max}

\def\Opt{{\tt Optimizer()}}

\ifthenelse{2>1}{ 
    \newcommand{\Juan}[1]{{\color{blue}Juan: #1}}
    \newcommand{\Michael}[1]{{\color{purple}Michael: #1}}
    \newcommand{\Juergen}[1]{{\color{ForestGreen}Juergen: #1}}
}{
    \newcommand{\Juan}[1]{}
    \newcommand{\Michael}[1]{}
    \newcommand{\Juergen}[1]{}
}

\title{Efficient computation of the\\ Knowledge Gradient for Bayesian Optimization}
\author{Juan Ungredda$^1$ \and Michael Pearce$^2$ \and Juergen Branke$^1$}
\date{$^1$University of Warwick, UK\\%
    $^2$Zenith AI, Opentrons\\Belfast, UK\\
September 2022}

\begin{document}

\maketitle

\begin{abstract}
    Bayesian optimization is a powerful collection of methods for optimizing stochastic
expensive black box functions. 
One key component of a Bayesian optimization algorithm is the acquisition function that determines which solution should be evaluated in every iteration.  A popular and very effective choice is the Knowledge Gradient acquisition function, however there is no analytical way to compute it. Several different implementations make different approximations.
In this paper, we review and compare the spectrum of Knowledge Gradient
implementations and propose One-shot Hybrid KG, a new approach that combines several of the previously proposed ideas and is  cheap
to compute as well as powerful and efficient.
We prove the new method preserves theoretical properties of previous methods 
and empirically show the drastically reduced computational overhead with equal or
improved performance. All experiments are implemented in BOTorch and code is
available on github.

\end{abstract}

\section{Introduction}
The problem of optimizing an expensive, stochastic, black box function appears in many domains, such as in simulation-based optimization \cite{ankenman2008stochastic}, in machine learning hyperparameter tuning \cite{snoek2012practical}, or in engineering optimization \cite{yamawaki2018multifunctional}.
In such
problems, the mapping between decision variables and outputs 
is not a simple mathematical expression but either a complex computer simulation, a wet lab biological experiment or machine learning training pipeline, from the perspective of an optimization algorithm, they are black boxes.
Formally, given a point in a low dimensional continuous
space(typically $D\leq 10$), $x\in X\subset \R^D$, we aim to find the point with the highest expected output
$$x^* = \argmax_{x\in X}\E[f(x)]$$
where $f:X\to R$ is a stochastic black box function and the expectation is over the stochasticity
in repeated calls to $f(x)$, e.g. multiple simulation runs with a different random number seed.

For such problems, Bayesian optimization (BO) methods have become a powerful and widely studied
toolbox for finding the optimum
using as few expensive black box evaluations as possible.
BO methods consist of two main components, a Gaussian Process surrogate model,
and an acquisition function. The surrogate model is trained to predict $f(x)$, outputting both a prediction
and an uncertainty/confidence. The acquisition function, $\alpha(x)$, quantifies the exploration, exploitation trade-off
for evaluating the black box at a new point $x$, and the new point with the highest acquisition value is then passed
to the black box for evaluation.

There exist many acquisition functions, the arguably most commonly used method is Expected Improvement (EI) \cite{jones1998efficient} that measures the expected amount that a new output $y=f(x)$ improves over the current best sampled output. \cite{srinivas2010} proposes to optimize an optimistic upper-confidence bound (UCB) acquisition function where the benefit of a point $x$ is quantified using a quantile of the distribution $f(x)$. EI and UCB are quick any easy to implement because there is an analytical solution. Thompson sampling (TS) \cite{Thompson1933} corresponds to using the GP to sample a set of predicted objective function values at a finite set of locations, the point with largest sample realization is chosen for evaluation, one may think of the sampled function as a randomly generated acquisition function to be maximized. This acquisition function is simple to implement though scales cubically with discretization size and more involved tricks are required otherwise. In contrast with EI and UCB, there exist many acquisition functions with more sophisticated theoretical motivation that also comes with much greater implementation difficulty. entropy Search (ES) \cite{hennig2012entropy} considers an information-based acquisition function where a GP over outputs generates a distribution over the input domain that models the probability of any input being the maximizer. This acquisition criterion involves computing the expected entropy reduction of the global solution and may be used for noisy observations. Computing the distribution over the input domain and its entropy and the same distribution and entropy for each possible future outcome, an average of one step look ahead entropies, introduces extensive mathematical and implementation problems to be tackled. Predictive Entropy Search \cite{ES3_hernandez2014predictive} proposed a quicker implementation however with more sophisticated approximations. Max Value entropy search \cite{wang2017max} methods aim to reduce entropy of the predictive distribution of the output value and cheaper methods have been proposed.

Knowledge Gradient (KG) \cite{frazier2009knowledge} is derived from Bayesian Decision theory
and samples locations that provide the greatest increase in the peak of of the GP posterior mean.
However, as with entropy methods, numerical approximations are required and several implementations using different approximations have been proposed in the literature varying in their complexity, accuracy and computational cost, we provide a detailed review of these methods in Section \ref{sec:kg_background}.
Some are simple to implement and find optimal points in simple convenient cases
while struggling in more challenging problems. Other KG implementations are more
involved to implement (open source versions are recommended) and also perform very
well across a broad range of use cases but incur a larger computational overhead.

With this work, we aim to take steps towards an algorithm that has both well founded
theoretical motivation (like KG and entropy methods) while also being easy to implement
and cheap to compute (like EI and UCB methods). We in particular focus on KG methods.
In this manuscript, we give a detailed review of the major technical milestones in
the development of KG implementations and then merge these enhancements to provide
an implementation of KG that is simple and cheap to compute, yet also practical and performing
well across a broad range of problems.

We hope this can make KG a far more accessible Bayesian Optimization acquisition
function for the average user and newcomers to the field, and due to its
computational efficiency also broaden the scope of problems for which KG is a preferred choice.


In Sections~\ref{sec:BO_background} and \ref{sec:kg_background}, we provide the mathematical
background on BO and the major milestones of KG implementation. In 
Section~\ref{sec:one-shot-hybrid-KG} we describe a natural novel
implementation, One-Shot-Hybrid KG, and discuss it's complexity,
theoretical properties and practical implementation. In Section~\ref{sec:numerical}
we present numerical ablation studies across the methods comparing time and opportunity
cost and finally conclude in Section~\ref{sec:conclusion}.

\section{Bayesian Optimization} \label{sec:BO_background}
Bayesian optimization sequentially collects data from the black box function and builds a surrogate model, most often a Gaussian process (GP)  model. Let the $n$ collected inputs be denoted $X^n=(x_1, x_2, \ldots, x_n) \subset X$ with outputs $Y^n\in\R^n$ and $\D^n$ the dataset of pairs. A Gaussian process is specified by prior mean, $\mu^0(x)$ that is typically constant $\mu^0(x) = \text{mean}(Y^n)$, and prior kernel  $k(x, x')\in \R$ that gives the covariance (expected similarity) between the output values at $x$ and $x'$. Common choices of kernel are the Squared Exponential (RBF) and Mat\'ern kernel
\begin{eqnarray*}
r_l^2 &=& \sum_{i=1}^D\frac{(x_i-x_i')^2}{l_i^2} \\
k_{RBF}(x, x'; \sigma_0, l) &=& \sigma_0\exp\left(-\frac{r_l^2}{2}\right) \\
k_{Mat}(x, x'; \sigma_0, l) &=& \sigma_0\left(1 + \sqrt{5}r_l + 5r_l^2/3 \right)\exp\left(-\sqrt{5}r_l\right)
\end{eqnarray*}
where the $\theta = \{\sigma_0, l_1,...,l_D\}$ are hyperparameters estimated
using maximum marginal likelihood \cite{rasmussen2003gaussian}.
In our experiments, we adopt the squared exponential kernel.
After observing $n$ points, the \emph{posterior} mean and covariance functions  are given by
\begin{eqnarray}
    \mu^n(x) &=& k(x, X^n)\big(k(X^n, X^n) + \sigma^2I\big)^{-1}Y^n,\\
    k^n(x, x') &=& k(x, x') - k(x, X^n)\big(k(X^n, X^n) + \sigma^2I\big)^{-1}k(X^n, x').
\end{eqnarray}
For each BO iteration, the latest data is used to build a new model, then given a new \emph{candidate} point $x\n1$, an acquisition function $\alpha(x\n1, \mu(\cdot), k(\cdot, \cdot))$ quantifies
the expected benefit of evaluating the black box at $x\n1$, accounting for both exploration and exploitation. The acquisition function is optimized over $X$ to find the most beneficial
next point ${x\n1}^*  =\argmax_{x\n1}\alpha(x\n1,\cdot)$ which is then passed to the
expensive black box function $y\n1 = f({x\n1}^*)$. The dataset is updated, $\D\n1$, and the next iteration starts, see pseudocode in Algorithm~\ref{alg:BO_framework}. In this work, we
focus exclusively on the Knowledge Gradient acquisition function and its many implementations.
 
\begin{algorithm}
    \label{alg:BO_framework}
    \caption{The Bayesian Optimization Algorithm. An initial dataset of $n_0$ points is collected
    over the domain $X$. Then new points are sequentially determined and evaluated for the rest
    of the budget $N$. In each round, a Gaussian process regression model is fit to the current dataset and the acquisition function is optimized to find the next point to evaluate. Finally, the best predicted point is returned.}
    \begin{algorithmic}
        \Require blackbox objective $f:X\to\R$, budget $N$, initialisation budget $n_0$, GP kernel $k(x, x'|\theta)$, acquisition function $\alpha(x, \mu(\cdot), k(\cdot, \cdot))$
        \State $X^{n_0}\gets \text{LHC}(X, n_0)$ \Comment{initial inputs, latin hypercube over $X$}
        \State $\D^{n_0} \gets \big\{(x^i, f(x^i)) \big| x^i \in X^{n_0}\big\}$ \Comment{make initial dataset}
        \For{$n = n_0,\dots, N-1$}
            \State $\mu^n(\cdot), k^n(\cdot, \cdot) \gets \GP(\D^n, k(\cdot, \cdot))$ \Comment{construct GP model}
            \State ${x\n1}^* \gets \argmax_{x\n1}\alpha(x\n1, \mu^n(\cdot), k^n(\cdot, \cdot))$ \Comment{optimize the acquisition function}
            \State $y\n1 \gets f({x\n1}^*)$ \Comment{evaluate black box at next point}
            \State $\D^{n+1}\gets \D^n\cup\{({x\n1}^*, y\n1)\}$ \Comment{update dataset}
        \EndFor
        \State $\mu^N(\cdot), k^N(\cdot, \cdot) \gets \GP(\D^N, k(\cdot, \cdot))$ \Comment{construct final GP model}
        \State \textbf{return} $x^* = \argmax_x \mu^N(x)$ \Comment{return best predicted point}
    \end{algorithmic}
\end{algorithm}


\section{A Tour of Knowledge Gradient Implementations} \label{sec:kg_background}
We aim to create a simple, easy to use Knowledge Gradient implementation. In the following, we provide a 
mathematical review of existing methods after which we present our new method as the
natural next step in Section~\ref{sec:one-shot-hybrid-KG}.

Given a set of past observations $\D^n$ and a proposed new sampling point
location $x\n1$, Knowledge Gradient (KG) quantifies the value of a new hypothetical
observation $y^{n+1}=f(x\n1)$ by the expected increase in the peak of the posterior mean
\begin{equation}\label{eq:KG_global}
    \KG(x\n1) = \E_{\yn1}\big[\max_{x'} \mun1(x')\big| x\n1\big] - \max_{x''\in X} \mu^n(x'').
\end{equation}
where we suppress arguments $\mu^n(\cdot)$, $k^n(\cdot, \cdot)$ and $\D^n$ for brevity. Unfortunately, $\max_{x'\in X} \mun1(x')$ and the enclosing expectation has no explicit formula
and approximations are required. We here emphasize, \emph{accurate approximation
is the central challenge in implementing Knowledge Gradient methods} and has
been the focus of many prior works. These methods rely on the following
``reparameterization trick": at time $n$, the new posterior mean is an unknown random function, however, it may be written as
\begin{equation}\label{eq:mu_n1_Z}
    \mu^{n+1}(x) = \mu^n(x) + \tilde\sigma(x; x\n1)Z
\end{equation}
where $\tilde\sigma: X \times X \to \R$ is a deterministic scalar valued function and the scalar random variable $Z\sim\N(0,1)$
captures the posterior predictive randomness of the yet unobserved $y\n1$, see Appendix~\ref{apndx:one_step_post_mean}. Hence one may also write
\begin{equation}\label{eq:KG_global_Z}
    \KG(x\n1) = \E_{Z}\big[\max_{x'} \mu^n(x') + \tilde\sigma(x'; x\n1)Z\big] - \max_{x''\in X} \mu^n(x'').
\end{equation}
Moreover, by Jensen's inequality and the convexity of the $\max()$, it is easily shown that $\KG(x\n1) \geq 0$, there is never an \emph{expected} disadvantage to collecting more data.


\subsection{Discrete Knowledge Gradient}
The early KG methods for continuous spaces, \cite{Frazier2009, scott2011correlated}, approximated $\KG(x\n1)$ by replacing the domain of the inner maximization from the continuous space $X$ to a finite discretization of $d$ points, $X_d\subset X$. $X_d$ may simply be a latin hypercube design over $X$, or the past sampled points $X^n$ or both. Denoting vectors $\underline{\mu}=\mu^n(X_d)\in\R^d$ and $\underline{\tilde\sigma}(x\n1) = \tilde\sigma(X_d; x\n1)\in\R^d$, then
$$
\KG_d(x\n1, X_d) = \E_Z\left[ \max \{\underline \mu + \underline{\tilde\sigma}(x\n1)Z \} \right] - \max \underline \mu.
$$
The $\max \{\underline \mu + \underline{\tilde\sigma}(x\n1)Z \}$ is a piece-wise
linear function of $Z$, thus the expectation over Gaussian $Z$, and therefore $\KG_d(\cdot)$, is analytically tractable, the algorithm has been proposed in \cite{Frazier2008} and is provided in Appendix~\ref{apndx:discrete_KG_epigraph} for completeness.
If the current best predicted point is in the discretization, $\argmax \mu^n(x) \in X^d$, the discrete Knowledge Gradient is a \emph{lower bound} of the true Knowledge Gradient
$$ 0 \leq \KG_d(x\n1, X_d) \leq \KG(x\n1)$$
and increasing the density of points in $X_d$ such that $X_d\to X$ tightens the bound.
The REVI \cite{pearce2018continuous} and the MiSo \cite{poloczek2017multi} algorithms
used $\KG_d(\cdot)$ with 3000 uniformly random distributed points.
While this method provides an analytic lower bound, it suffers the curse of
dimensionality. To be space filling, the number of points in $X_d$ must grow
exponentially with input dimension $D$. Further, a totally random discretization is highly likely to contain many useless points in uneventful regions of the space $X$ resulting in wasted computation, a sparse $X_d$ results in a loose ineffective lower bound, see Figure~\ref{fig:KG_demo} centre-left plot.


\subsection{Monte-Carlo Knowledge Gradient}

\begin{figure*}[t]
    \centering
    \includegraphics[width=0.98\textwidth]{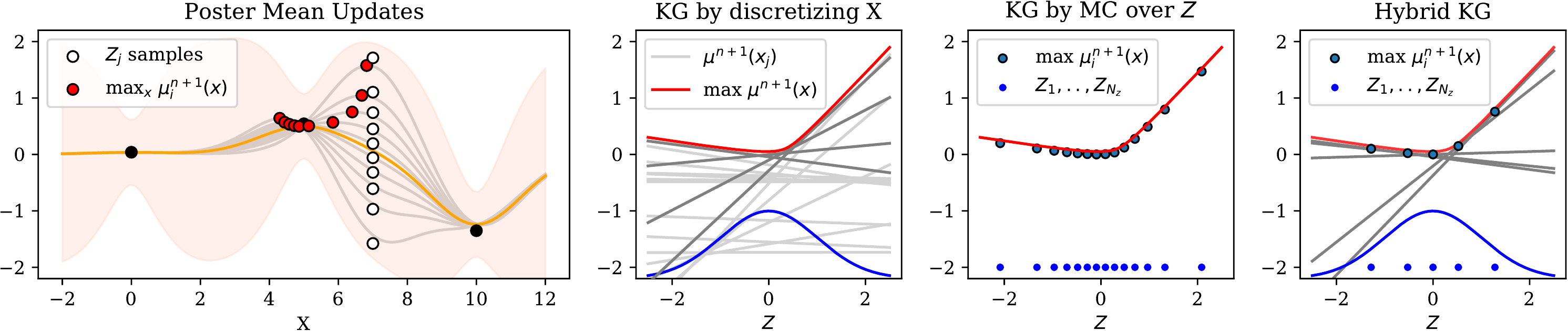}
    \caption{
    \label{fig:KG_demo}
    Methods for computing $\KG(x\n1)$ at $x\n1=7$.
    Left: $\mu^n(x)$ and samples of $\mu\n1(x)$ determined by a scalar $Z\sim N(0,1)$.
    Centre-left: $\KG_d$ replaces $X$ with up to 3000 points $x_i\in X_d$ and $\mu\n1(x_i)$ is linear in $Z$.
    Centre-right: $\KG_{MC}$ samples up to 1000 functions $\mu\n1(x)$ functions and maximises each of them numerically.
    Right: $\KG_h$ samples up to 5 functions $\mu\n1(x)$ and maximizes them numerically, 
    the $\argmax$ points $x^*_1,..,x^*_5$ are used as $X_d$ in $\KG_d$.
    }
     \vspace{-0.3cm}
\end{figure*}

To avoid the curse of dimensionality, the expectation over $Z$ in
Equation~\ref{eq:KG_global_Z} may be stochastically approximated by Monte-Carlo sampling \cite{wu2017discretization,wu2017bayesian}. Given $x\n1$, the method samples $n_z$ standard Gaussian values, $Z_{MC}\in\R^{n_z}$.
For each sample, $Z_j \in Z_{MC}$, it constructs a corresponding posterior mean realisation, 
$$\mun1_j(x) = \mu^n(x) + \tilde\sigma(x; x\n1)Z_j,$$
and finds the maximum with a continuous numerical \Opt like L-BFGS \cite{liu1989limited} or conjugate gradient \cite{shewchuk1994introduction} with multiple restarts.
We use \Opt to denote a functional taking an arbitrary function $g:X\to \R$ as input and returning $\max_{x\in X} g(x)$ as output. The Monte-Carlo KG is then defined as the average of the maxima from all $Z_j$ as follows
\begin{align*}
\KG_{MC}(x\n1, Z_{MC}) = 
\frac{1}{n_s}\sum_j \underset{x'}{\text{\tt{Optimizer}}}\big(\mun1_j(x')\big) 
- \underset{x''}{\tt{Optimizer}}\big(\mu^n(x'')\big).
\end{align*}
Assuming {\tt Optimizer()} converges, the result is an unbiased, consistent stochastic estimate of true Knowledge Gradient. Slightly abusing $Z_{MC}$ notation, we have
\begin{eqnarray}
    \E_{Z_{MC}|n_z}\left[\KG_{MC}(x\n1,Z_{MC})\right] = \KG(x\n1), \\
    \lim_{n_z\to \infty} \KG_{MC}(x\n1,Z_{MC})  = \KG(x\n1).
\end{eqnarray}
For larger input dimension $D$, the \Opt over $X\subset \R^D$ may simply be run for more steps (linear in $D$) to converge thus avoiding the curse of dimensionality. Compared with discrete KG that discretizes optimization over $X$ and continuously integrates over (1 dimensional) $Z$, Monte-Carlo KG instead continuously optimizes over $X$ and discretely integrates over $Z$ with Monte Carlo samples, see Figure~\ref{fig:KG_demo} centre right.
However, for a good estimate, $n_z$ must be large, e.g. $n_z=1000$, and many \Opt calls are required. Furthermore, if $Z_j\approx Z_j'$, the optimal value may be near identical and need not be called twice. Finally to optimize $\KG_{MC}(x\n1\ Z_{MC})$ over $x\n1$, a stochastic gradient ascent optimizer is required, e.g. Adam \cite{kingma2014adam}, and  must be set up correctly to ensure convergence. A small choice of $n_z$ or poor inner \Opt increases bias and variance in the KG estimate.  Further, repeated calls to $\KG_{MC}(\cdot)$ for different values of $x\n1$ can be expensive as all the $Z_{MC}$ values are resampled and the \Opt calls must be executed from scratch.


\subsection{Hybrid Knowledge Gradient}
The Hybrid Knowledge Gradient first proposed in \cite{pearce2020practical} aims to combine the best of both Discrete KG (analytic tractability, speed) and Monte-Carlo KG (scalabilty to higher input dimensions). Given $x\n1$, a set of $n_z=5$ unique, deterministic $Z$ values is constructed
from uniformly spaced Gaussian quantiles 
$$Z_h = \{\Phi^{-1}(0.1), \Phi^{-1}(0.3), \Phi^{-1}(0.5), \Phi^{-1}(0.7),\Phi^{-1}(0.9)\} \subset \R$$
where $\Phi^{-1}:[0, 1]\to \R$ is the inverse Gaussian cumulative distribution function. Following Monte-Carlo KG, for each $Z_j\in Z_h$, the posterior mean realisation is constructed, $\mu\n1_j(x)$, and optimized with {\tt Optimizer()} however the resulting optimal input $x^*_j$ is stored in a set
$X_{MC}$, 
\begin{eqnarray}
    X_{MC} &=& \big\{x^*_j \big| \mu\n1_j(x^*_j) = \text{{\tt Optimizer}}(\mu\n1_j(x')), j \in {1,\dots,n_z}\big\}.
\end{eqnarray}
Finally, following Discrete Knowledge Gradient, the optimal inputs $X_{MC}$ form the discretization used in Hybrid KG, i.e.,
\begin{eqnarray}
    \KG_h(x\n1) &=& \KG_d(x\n1, X_{MC}).
\end{eqnarray}
Thus Hybrid Knowledge Gradient is a deterministic, analytic, \emph{maximized} lower bound to the true Knowledge Gradient. Hybrid KG scales to higher dimensional inputs like Monte-Carlo KG, while reducing computation using only $n_z=5$. Similar to Monte-Carlo KG, repeated calls to $\KG_h(x\n1)$ for different $x\n1$ still require executing all the {\tt Optimizer()} calls from scratch.


\subsection{One-Shot Knowledge Gradient}
With the goal of reducing the computation of Monte-Carlo KG, with a few changes, we next show how to derive One Shot KG \cite{balandat2020botorch}.
If we assume we are given a set of $Z_{MC}$ each with corresponding optimal points $X_{MC}$,
each $Z_j\in Z_{MC}$ is paired with a $x_j^* \in X_{MC}$,
the One-Shot estimate of KG is as follows, 
\begin{eqnarray}
    \KGOS (x\n1, Z_{MC}, X_{MC}) &=&  \frac{1}{n_z}\sum_j \mu\n1_j(x^*_j) - \max_{x'}\mu^n(x').
\end{eqnarray}
$\KGOS$ would be a very poor under estimate if $x^*_j\in X_{MC}$ points are random, and when the points are all optimized it recovers $\KGMC$.
In One-Shot KG, the random samples $Z_{MC}$ are fixed for each BO iteration hence $\KGOS$ is deterministic. Next, in the search for $x\n1$, we may \emph{simultaneously} search over $X_{MC}$
hence the $\KGOS$ estimate improves over the course of the search for the next candidate point $x\n1$,
\begin{eqnarray}
    {x\n1}^* &=& \argmax_{x\n1} \max_{X_{MC}} \KGOS(x\n1, X_{MC}, Z_{MC}).
\end{eqnarray}
equivalently, this acquisition function may be optimized with the same deterministic optimizer  
\begin{equation}
    \underset{x\n1,X_{MC}}{\tt Optimizer}\big(\KGOS(x\n1, X_{MC}, Z_{MC})\big),
\end{equation}
where $Z_{MC}$ are frozen constant values and all the $x$ points are optimized over the same domain $X$, the final optimal ${x\n1}^*$ is used as the next sample (the final optimized $X_{MC}$ is no longer explicitly required).
In Monte-Carlo KG and Hybrid KG, one optimizer searches for $x\n1$, and at each candidate $x\n1$, nested optimizers are applied to find $X_{MC}$, even if subsequent $x\n1$ are very close and optimization of $X_{MC}$ may be largely repeated. One-Shot KG optimizes both $x\n1$ and $X_{MC}$ at the same time in a single optimizer, significantly reducing computational
effort to find $X_{MC}$. However, freezing $Z_{MC}$ and not ensuring $X_{MC}$ is fully converged introduces bias.

\begin{figure}
    \centering
    \begin{tabular}{ccc}
    	\includegraphics[height=3.5cm]{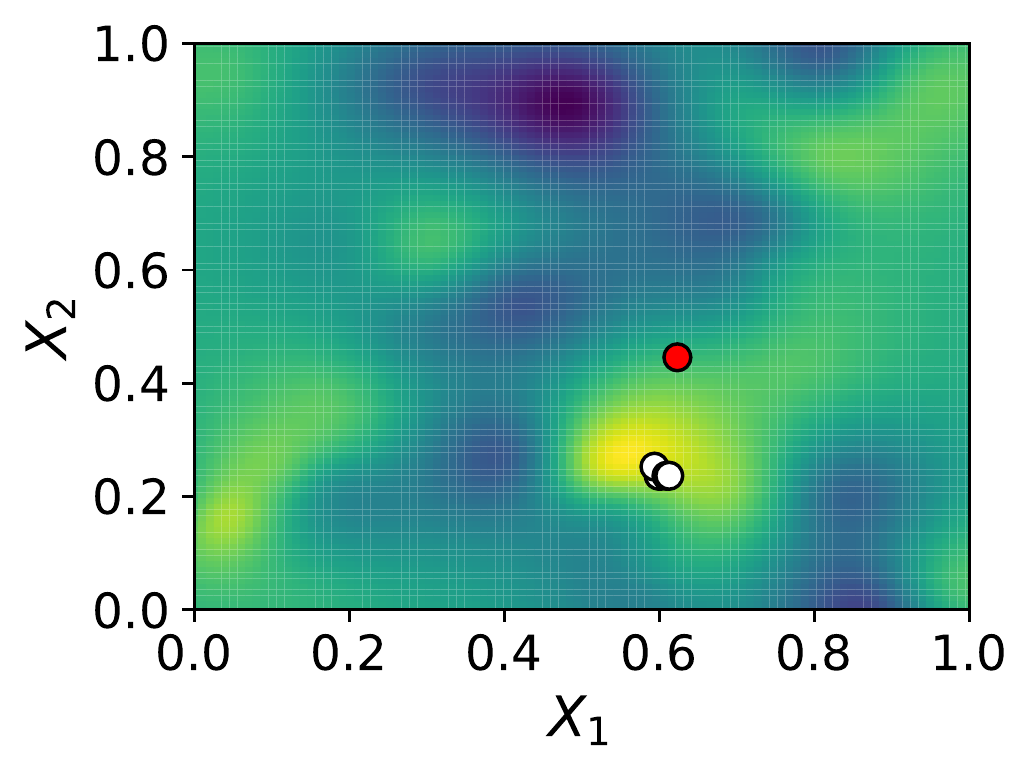}&
    	\includegraphics[height=3.5cm]{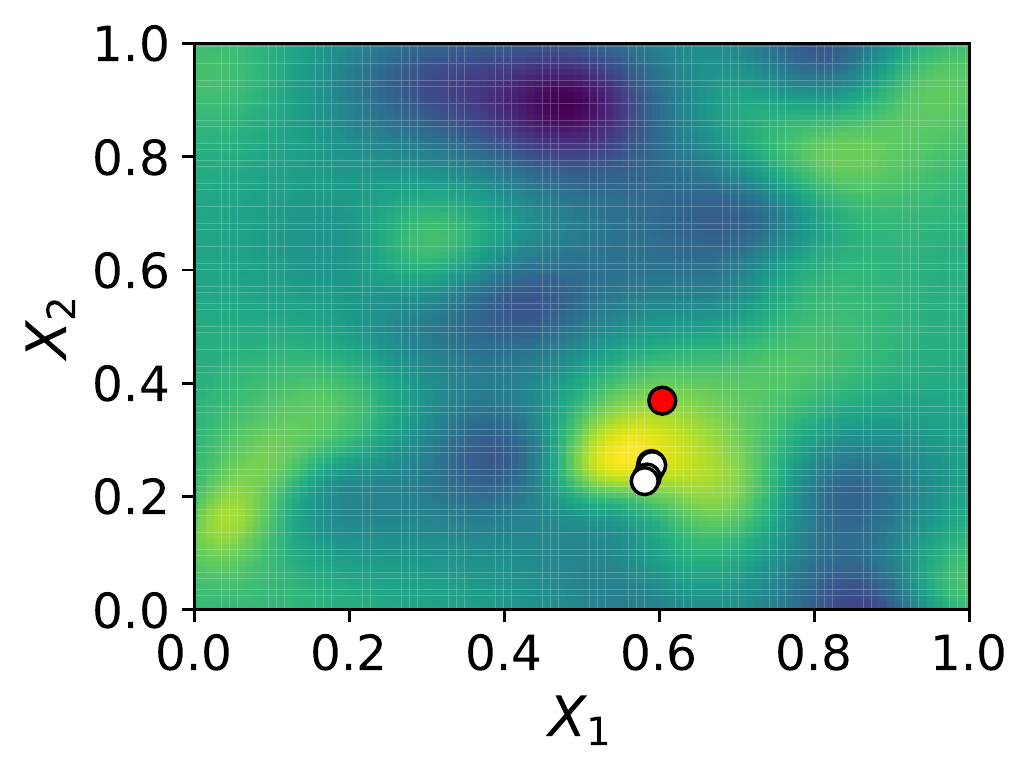}&
    	\includegraphics[height=3.5cm]{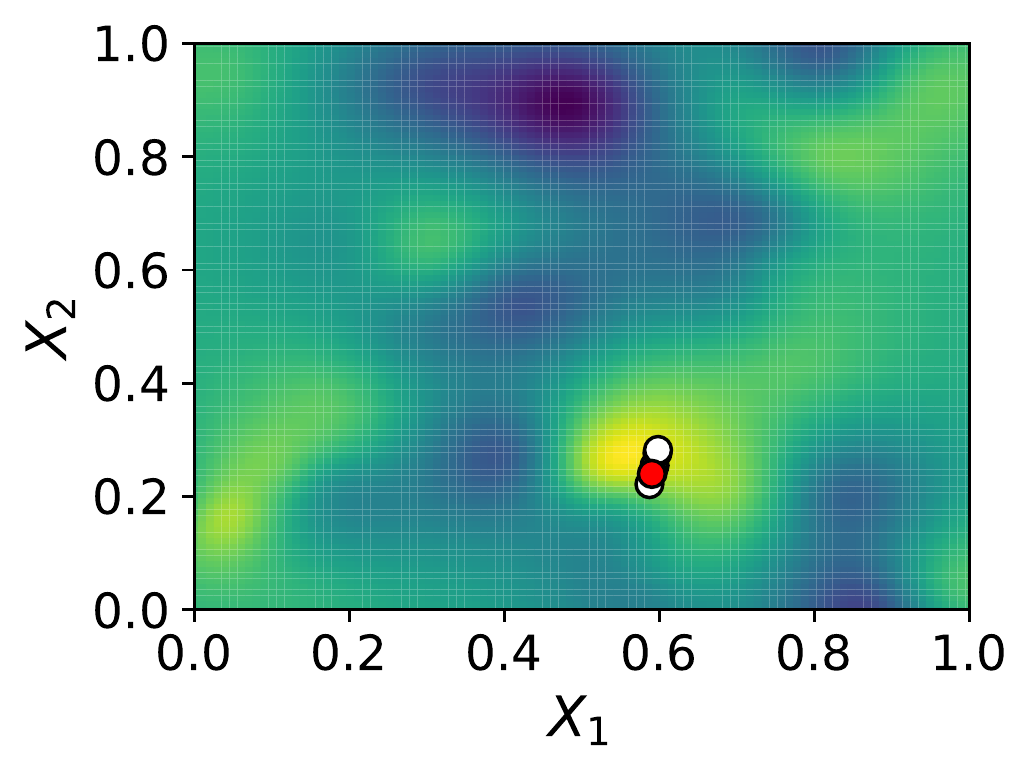}\\
    	(a) & (b) & (c)\\
    	
    	  \includegraphics[height=3.5cm]{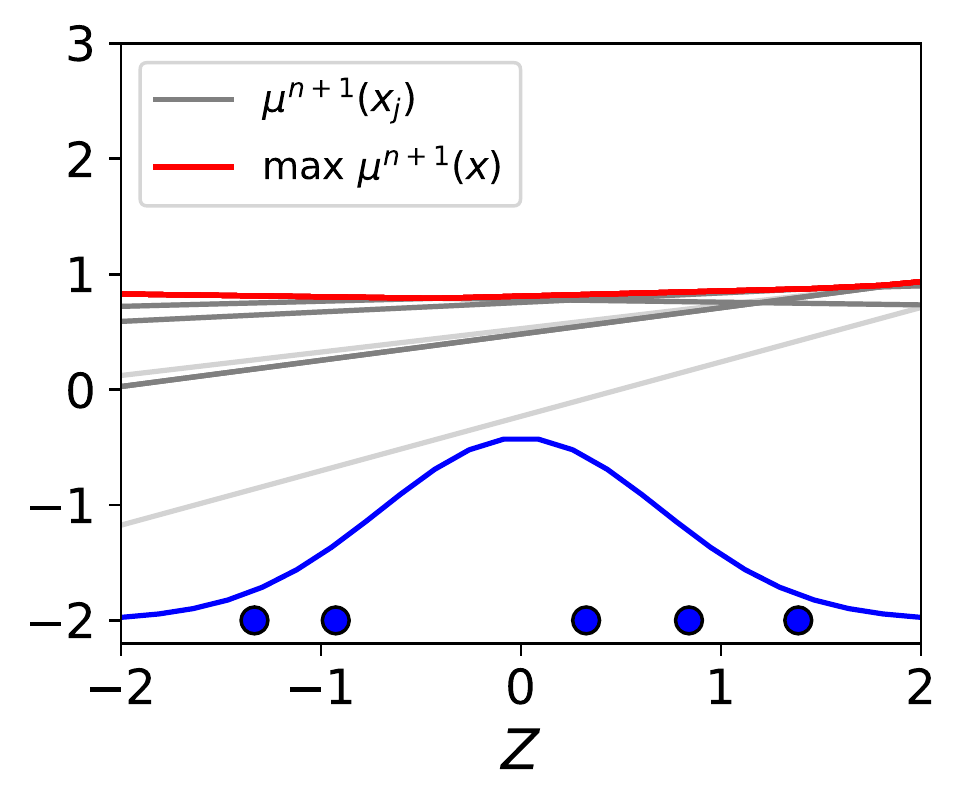}&
    	\includegraphics[height=3.5cm]{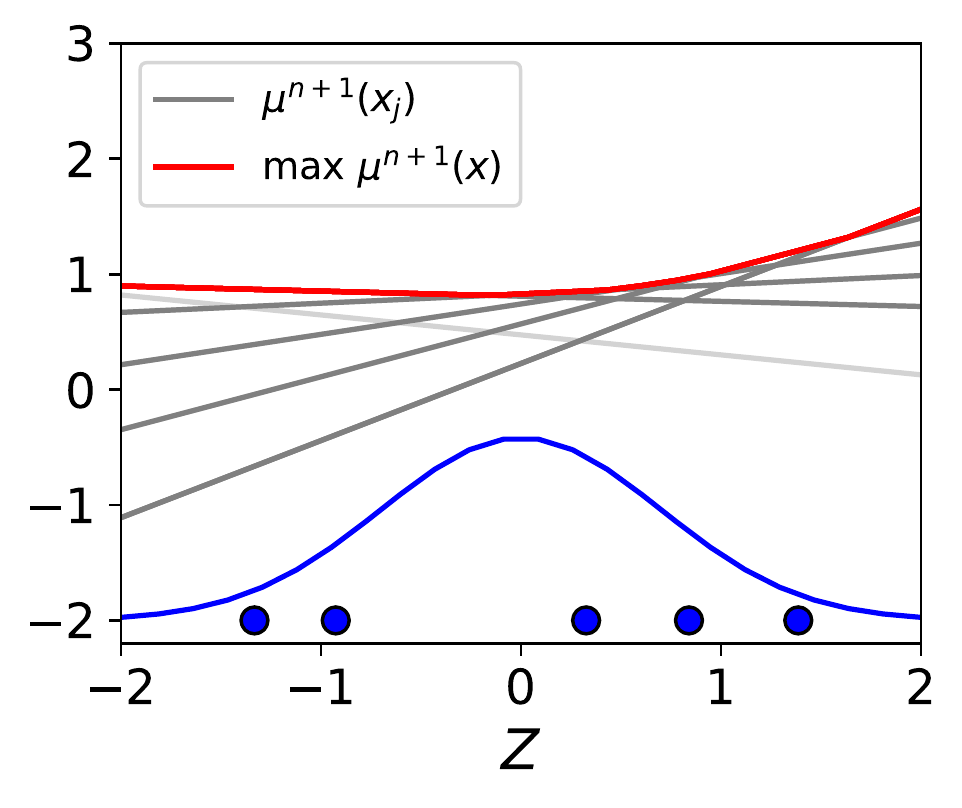}&
    	\includegraphics[height=3.5cm]{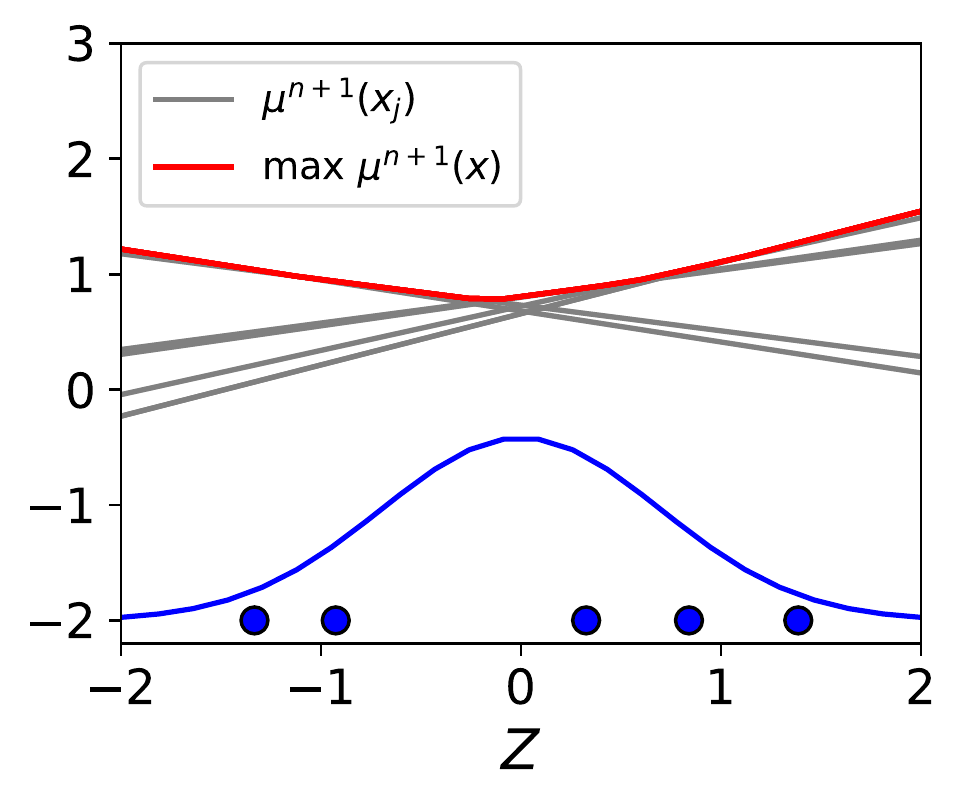}\\
    	(d) & (e) & (f)\\
    \end{tabular}

    \caption{Illustration of the $\KGOS$ acquisition function optimization.  (a) shows an initial sample $x^{n+1}$ (red) and set $X_{MC}$ over a black-box function landscape, with brighter colors indicating higher function values. (b) shows the resulting $x^{n+1}$ and $X_{MC}$ after applying \Opt to the acquisition function where both, $x^{+1}$ and $X_{MC}$ are optimized at the same time in the optimizer. (c) shows the final $x^{n+1^*}$ and $X_{MC}$ achieved by the optimizer.(d-f) shows the surface of the maximum posterior over the set $X_{MC}$ given the values of $Z_{MC}$ (blue dots). }
\end{figure}

\section{One Shot Hybrid Knowledge Gradient} \label{sec:one-shot-hybrid-KG}
In this work we propose a simple unification of the aforementioned innovations. We take discrete KG and and make the discretization an explicit variable to be optimized along with the next sample point, that is 
\begin{eqnarray}
    {x\n1}^* = \argmax_{x\n1}\max_{X_d} \KG_{OSH}(x\n1, X_d)
\end{eqnarray}
where $\KG_{OSH()} = \KG_d()$ but we use separate notation for clarity here. The optimization is performed over the joint domain $(x\n1, X_d)\in X^{1+n_z}$. Note that neither $Z_h$ or $Z_{MC}$ are required. This method may be viewed as Discrete KG and One-Shot KG where both tricks have been applied simultaneously, the hybrid trick: enabling $n_z=|X_d|=5$ and a tight lower bound estimate of true KG, and the one-shot trick: simultaneous optimization drastically reducing execution time. As discrete KG is analytically tractable, the gradients with respect to both arguments are also analytically tractable and hence may be optimized with any \emph{deterministic} gradient ascent algorithm.

For a given discretization size, the One-Shot Hybrid KG has almost exactly the same computational cost as discrete KG. Both methods compute $\KG_d()$ and $\nabla_{x\n1}\KG_d()$ for gradient ascent over $x\n1$. However, One-Shot Hybrid KG also computes $\nabla_{X_d}\KG_d()$ for gradient
ascent over $X_d$. In practice we use PyTorch that supports automatic
differentiation via the back-propagation algorithm and GPU acceleration.

\begin{figure}
    \centering
    \begin{tabular}{ccc}
    	\includegraphics[height=3.7cm]{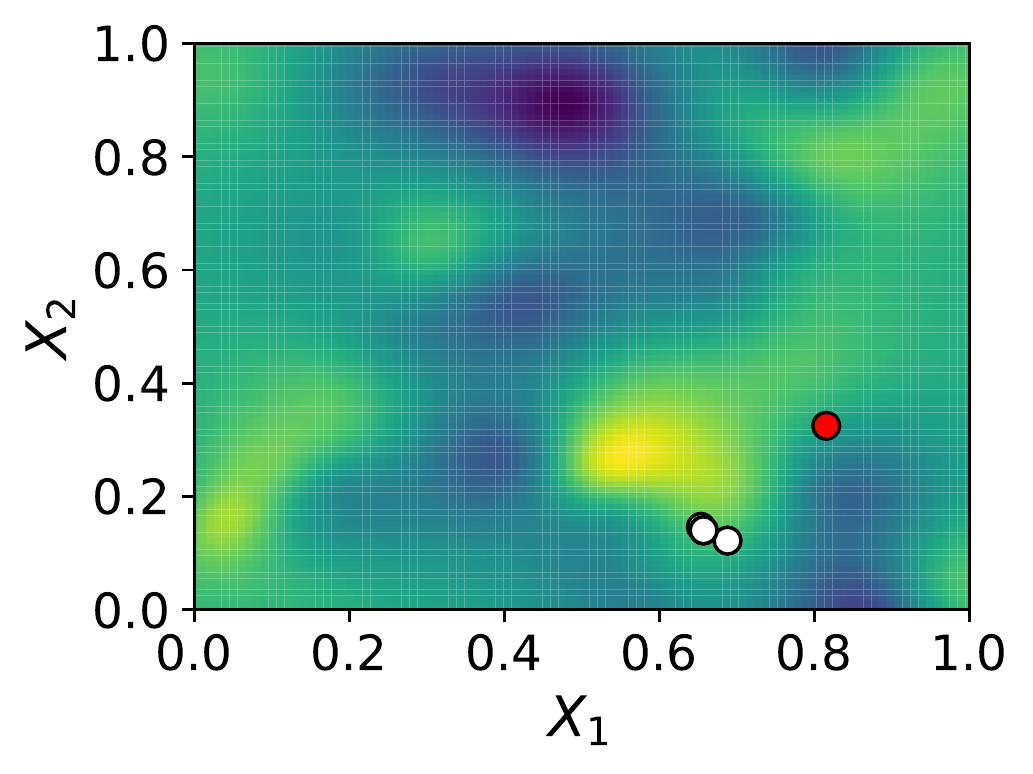}&
    	\includegraphics[height=3.7cm]{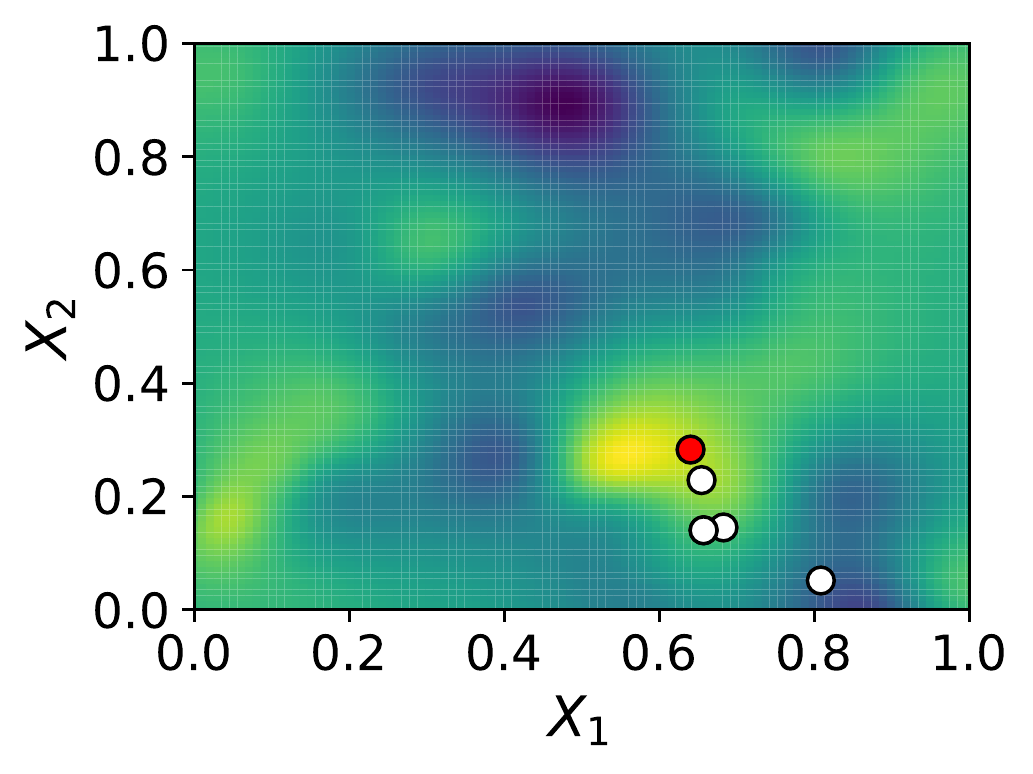}&
    	\includegraphics[height=3.7cm]{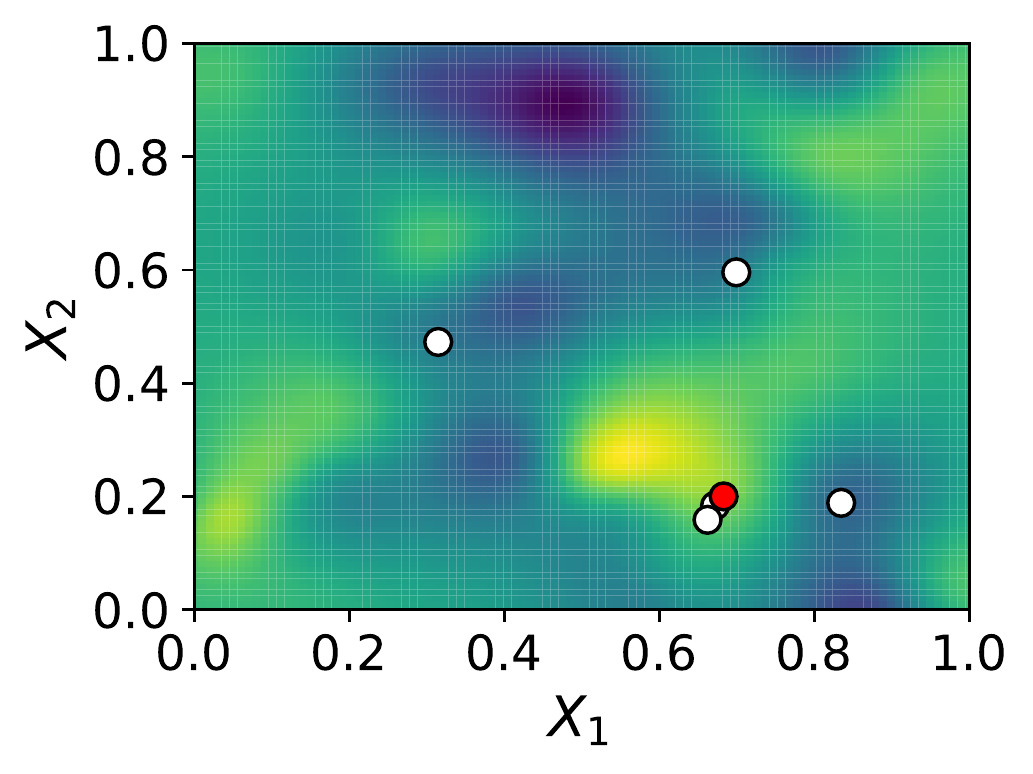}\\
    	(a) & (b) & (c)\\
    	
    	\includegraphics[height=3.7cm]{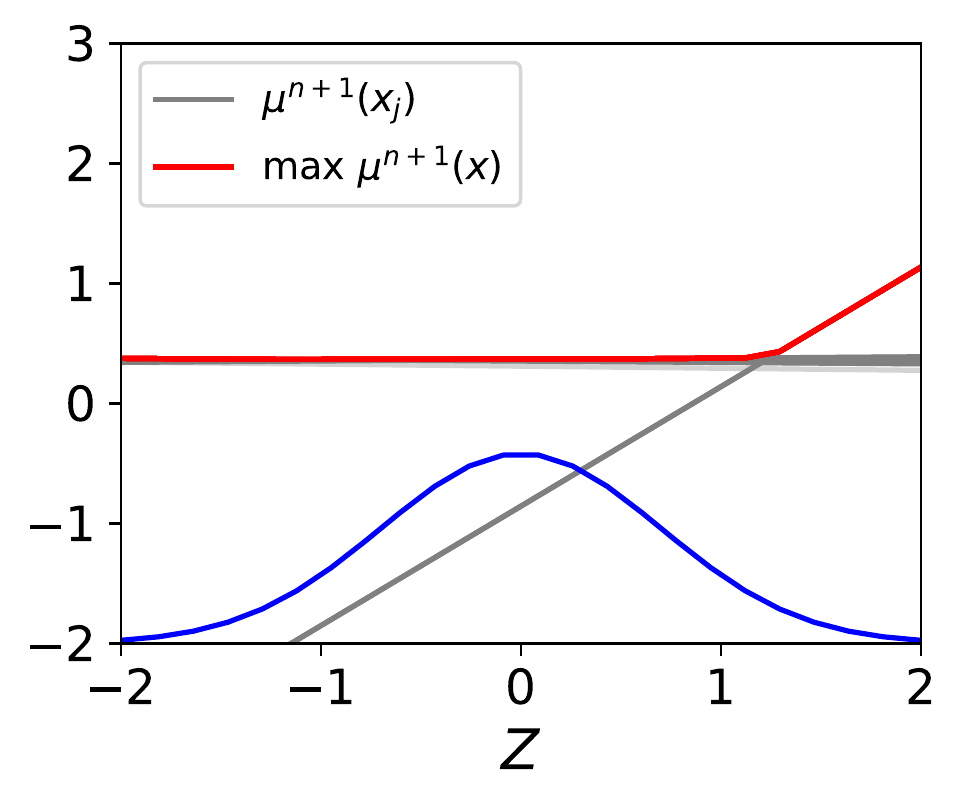}&
    	\includegraphics[height=3.7cm]{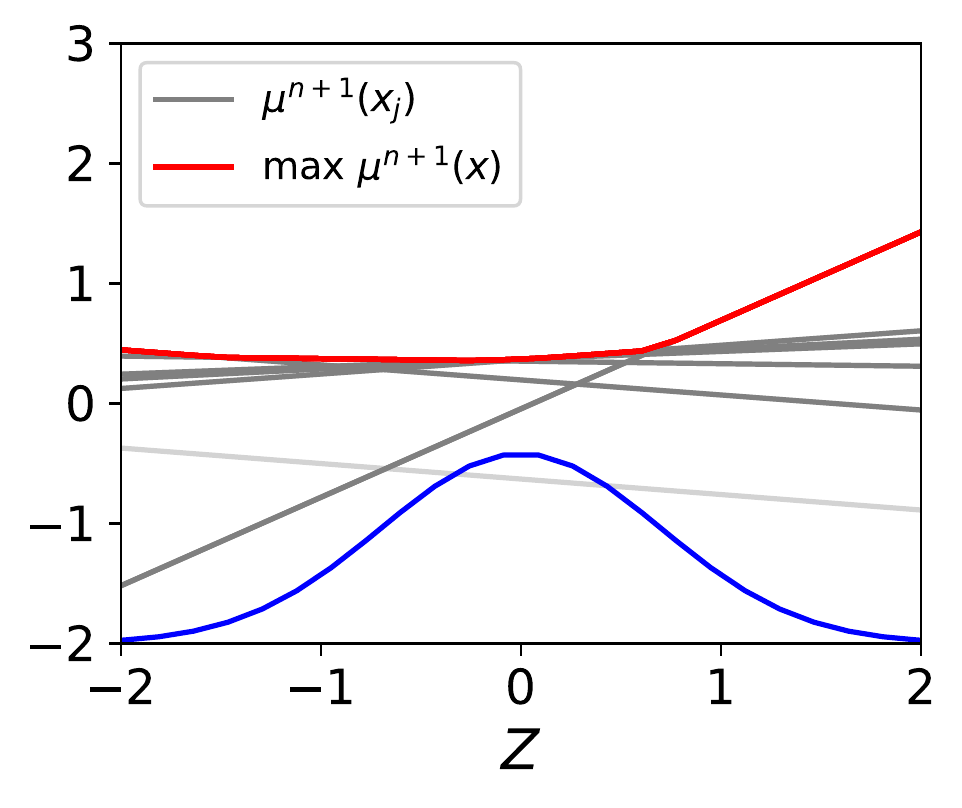}&
    	\includegraphics[height=3.7cm]{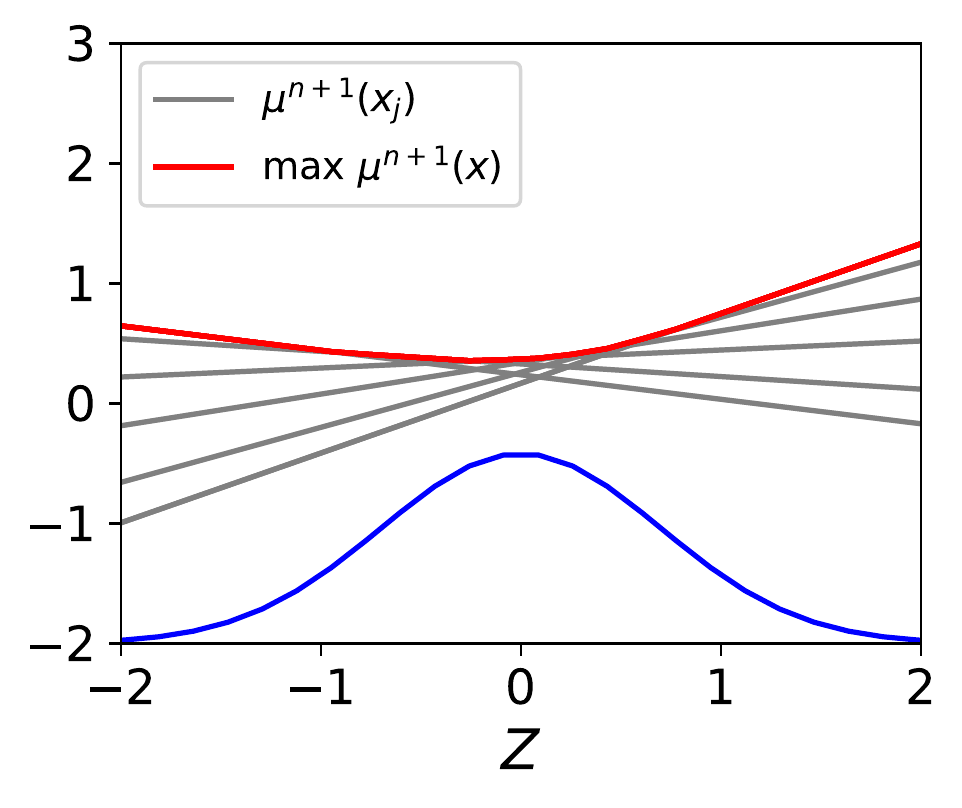}\\
    	(d) & (e) & (f)\\
    \end{tabular}

    \caption{Illustration of the One-Shot Hybrid KG acquisition function optimization.  (a) shows an initial sample $x^{n+1}$ (red) and set $X_{MC}$ over a black-box function landscape, with brighter colors indicating higher function values. (d) shows the surface of the maximum posterior over the set $X_{MC}$ for $Z$. The One-Shot Hybrid KG aims to maximize the expectation of the piece-wise linear function (red). (b) shows the resulting $x^{n+1}$ and $X_{MC}$ after applying \Opt to Discrete KG where both, $x^{+1}$ and $X_{MC}$ are optimized at the same time in the optimizer. (c) shows the final $x^{n+1^*}$ and $X_{MC}$ achieved by the optimizer with an optimized epigraph (f). }
\end{figure}

\subsection{Theoretical Properties}
As Hybrid One-Shot KG is simply an extension of Discrete KG, it inherits the theoretical properties of Discrete
KG in continuous spaces previously proven in \cite{Scott2011a}. The algorithm converges in the limit of infinite
budget, with infinitely many BO iterations and calls to the expensive black box, the true optimal input will be found. 
We only require that $\KG_d(x\n1, X_d) \geq 0$ for all $x\n1 \in X$, which is trivially
satisfied by enforcing that $x^*_n = \argmax \mu^n(x)$ is included in the set $X_{MC}$
and thus
\begin{eqnarray}
    \max_{X_{d}}\KG_d(x\n1, X_{d})&=& \max_{X_{d}} \E_Z\left[\max \mu^n(X_{d}\cup\{x_n^*\}) + Z \tilde\sigma(x\n1, X_{d}\cup\{x_n^*\})\right] - \max\mu^n(x) \\
    &\geq& \E_Z\left[\mu^n(x^*_n) + Z \tilde\sigma(x\n1, x^*_n)\right] - \max \mu^n(x) \\
    &=& \mu^n(x^*_n) + \E[Z]\tilde\sigma(x\n1, x^*_n) - \max \mu^n(x) \\
    &=& 0.
\end{eqnarray}
The equality holds when $k^n(x\n1, x) = c$ (typically $c=0$), and there is no benefit in
sampling $x\n1$, if this equality holds for all $x\n1\in X$, it can be shown that the true optimal
input is known. Further details can be found in \cite{poloczek2017multi, pearce2022bayesian}.


Also inherited from Discrete KG is the consistency of the One Shot Hybrid KG estimator as the discretization size increases to infinity,
increasing discretization size increases accuracy of the KG estimate.
Let $X_d^k = \{x_i|x_i\sim U(X), i=1,....,d\}$ be the uniformly randomly
generated discretization over $X$ with $d$ points, then we have that
\begin{eqnarray}
    \lim_{d\to\infty} \KG_{OSH}(x\n1, X_d) = \lim_{d\to\infty} \KG_{d}(x\n1, X_d) = \KG(x\n1).
\end{eqnarray}
While the result may be clear, the practical implication of this is two fold. Firstly, $d$ is an algorithm hyperparameter.
One may choose to increase $d$ and improve the accuracy (and cost) of each $\KG_{OSH}()$ call or alternatively, one may run
$\Opt$ for more iterations so that even for small $d$ the sparse $X_d$ converges towards an optimum.
In contrast, for one-shot KG where the \emph{first} call to $\KG_{OS}()$ with random $X_{MC}$ is a poor estimate of true KG
regardless of $n_z$, increasing the hyperparameter will not increase KG estimate accuracy, the algorithm requires \Opt~to be
run for multiple iterations for the KG estimate to become more accurate.
Hence One Shot hybrid KG may be somewhat less sensitive to hyperparameter settings.
In our experiments, we run the methods for a range of hyper parameter settings
comparing final performance however creating a strictly controlled experiment for comparison is a non trivial task which
we leave to future work.

\section{Numerical Experiments}\label{sec:numerical}

In this section we compare all KG implementations under the following acquisition function parameters:

\begin{itemize}
    \item Discrete Knowledge Gradient (DISC): We test this approach under 3, 10, and 1000 quasi-random uniformly distributed points.
    \item Monte-Carlo Knowledge Gradient (MC): We generate $n_{z}=$ 3 and 10 quasi-random standard Gaussian values.
    \item Hybrid Knowledge Gradient (HYBRID):  We generate $n_{z}=$ 3 and 10 uniformly spaced Gaussian quantiles.
    \item One-Shot Knowledge Gradient (ONESHOT): We generate $n_{z}=$ 3, 10, 128, and 500 quasi-random standard Gaussian values.
    \item One-Shot Hybrid Knowledge Gradient (ONESHOT-HYBRID): We optimize over a discretization size of 3 and 10.
\end{itemize}

For each method that depends on quasi-random samples, we fix the samples at each BO iteration. The resulting acquisition function is an entirely deterministic optimization problem and may be optimized using a deterministic optimizer. For One-Shot Knowledge Gradient, we used implementations available in BOTorch \cite{balandat2020botorch}. The remaining algorithms have been implemented from scratch.

\subsection{GP-Generated Experiments}

We consider a 100 test functions generated from a Gaussian process with a squared exponential kernel and hyper-parameters $l_{X} = 0.1$, $\sigma^2_{0}=1$. All functions are generated on a continuous space $X = [0,1]^{D}$ with dimensionality $D=\{2,6\}$, and without observation noise. The total budget of evaluations is set to $B=100$ and the results over the 100 test functions are aggregated to obtain confidence intervals (CI). To obtain the wall clock time, we measure the acquisition function evaluation time of each generated test function immediately after the initial design is evaluated.

We initially train the Gaussian process model to a set of $2(D + 1)$ initial black-box evaluations from the overall budget using a Latin hypercube (LHS) ‘space-filling’ experimental design. Furthermore, we assume that the hyper-parameters are known throughout the whole run of the algorithm to avoid the issue of model mismatch.

\begin{figure}
    \centering
    \begin{tabular}{cc}
    Dim: 2 & Dim: 6 \\
    	\includegraphics[height=5cm]{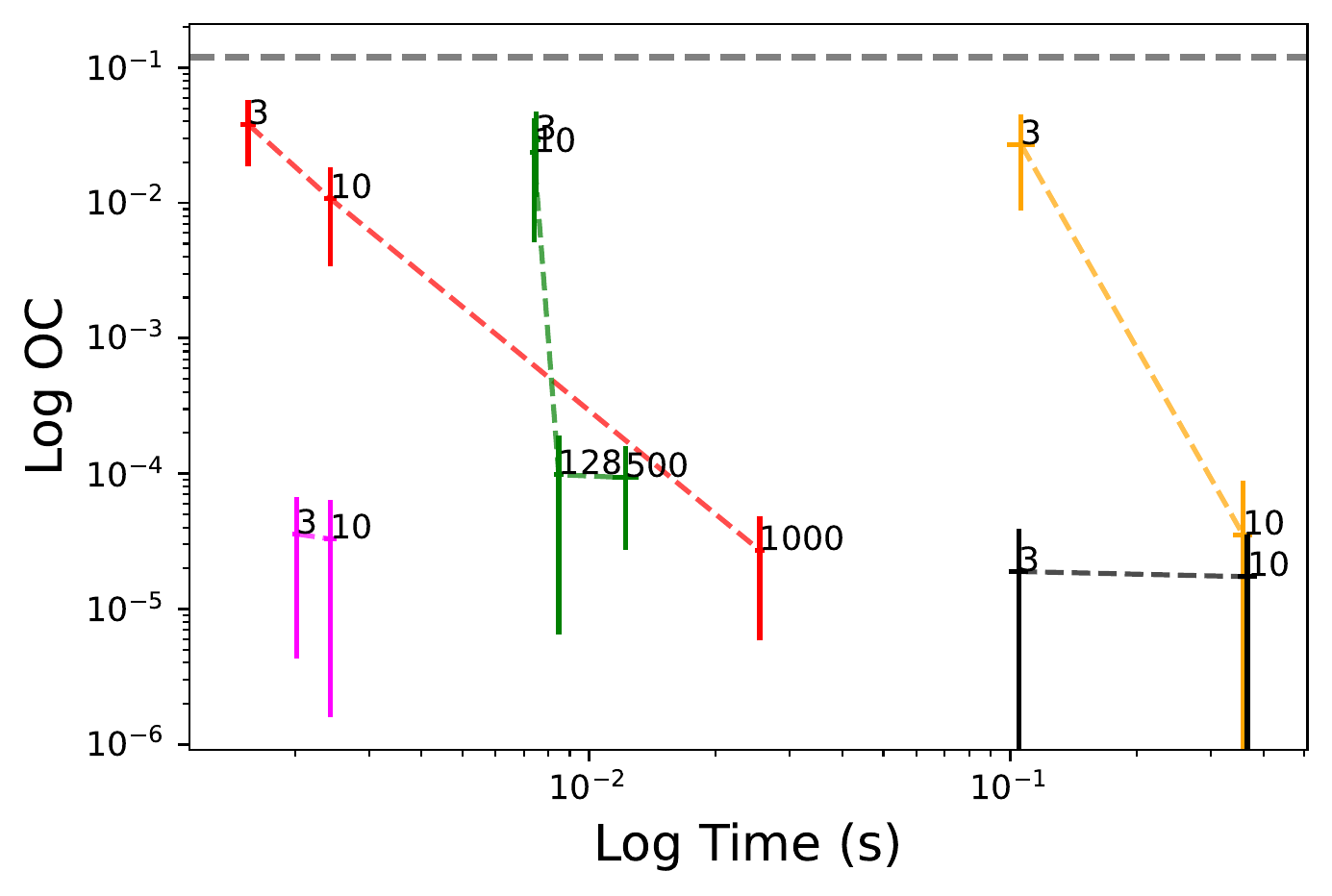}&
    	\includegraphics[height=5cm]{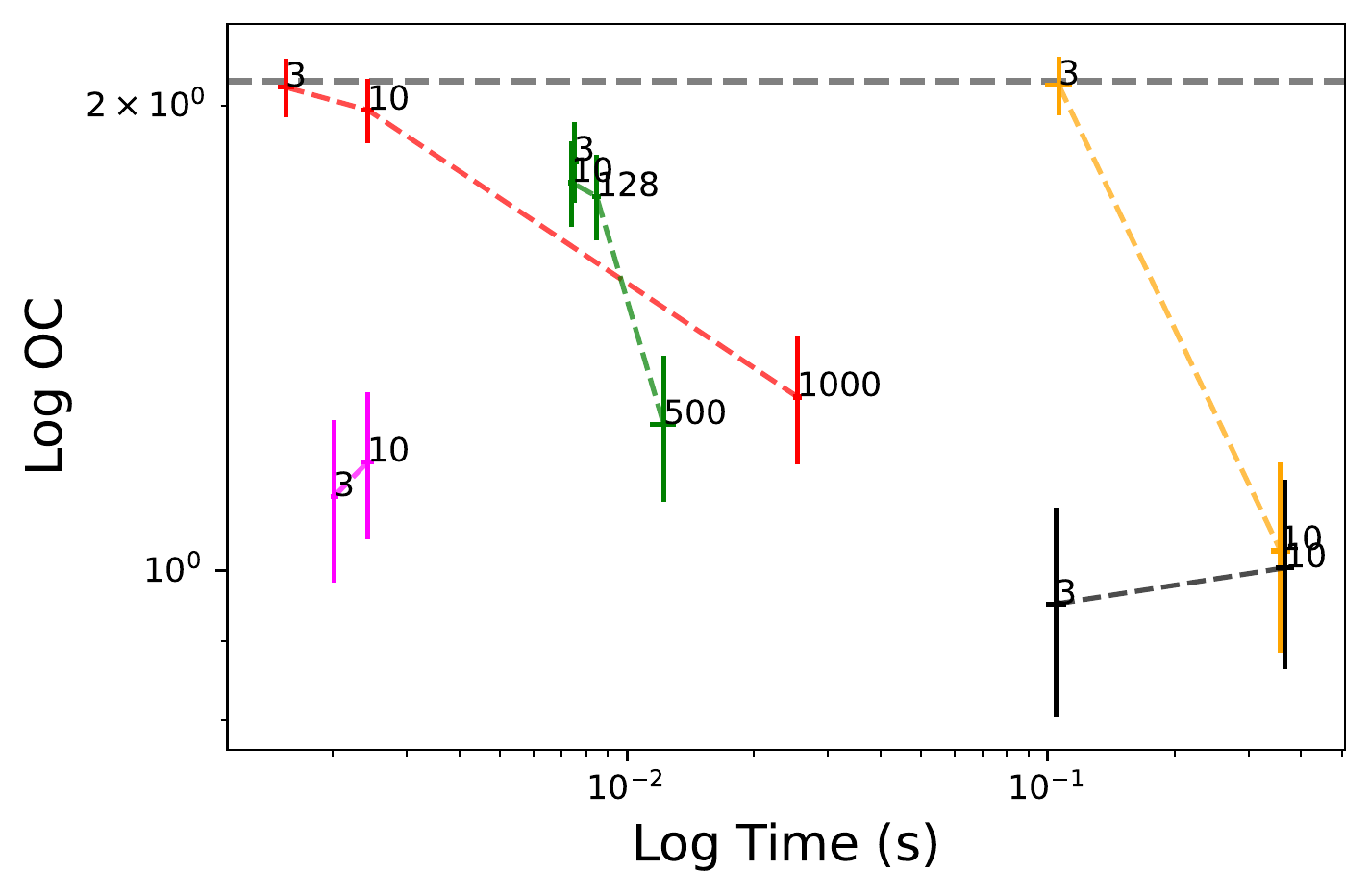}\\
    	(a) & (b)\\
    \end{tabular}
    
    \begin{tabular}{c}
    	\includegraphics[height=0.7cm]{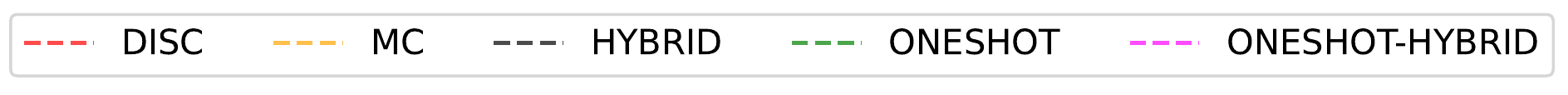}\\
    \end{tabular}
    \caption{Final Log OC vs Log Wall Clock Time (seconds) for (a) 2 design dimensions and (b) 6 design dimensions. In both plots, the mean performance of random sampling is shown as a grey horizontal line. All results are averaged over 100 independent test functions and both figures show the mean and 95\% confidence intervals for the OC.}
    \label{fig:OC_vs_ClockTime}
\end{figure}

Fig.~\ref{fig:OC_vs_ClockTime} shows the Opportunity cost (OC) once the budget, $B$, is depleted and the evaluation time in logarithmic scale. In both figures, DISC presents a performance close to random sampling when sparse discretizations are employed. Compared to other methods, a moderately high discretization size (1000) must be used to obtain competitive results. Notably, MC avoids the curse of dimensionality and drastically reduces the discretization size required compared to DISC. However, a small discretization size ($n_{z} =3$) produces high variance estimates of KG which reduces its performance. Furthermore, optimizing the discretization requires solving $n_{z}$ sequential inner optimization problems at each acquisition function  call which drastically increases the wall-clock time.

The HYBRID approximation improves over MC by generating a low variance approximation of KG which results in a superior performance when a low discretization is considered. However, HYBRID shows a similar evaluation time given by solving all inner optimization problems sequentially. On the other hand, ONESHOT avoids this problem by jointly optimizing the discretization space and the new solution. This results in a considerable decrease of the acquisition evaluation time, however, similar to DISC, ONESHOT relies on a moderately high discretization size to achieve competitive results. Lastly, the newly propopsed ONESHOT-HYBRID achieves a computational time comparable with DISC with competitive performance for low a higher discretization sizes.

\section{Conclusion}\label{sec:conclusion}

In this paper we considered the problem of implementing a fast and accurate approximation of KG. We proposed One-Shot Hybrid Knowledge Gradient, a fast method to compute KG that scales to higher dimensions. We empirically demonstrate the effectiveness of the proposed approach where One-Shot Hybrid Knowledge Gradient is both fast to compute and preserves its performance even under low discretization sizes in higher dimensions.

As future work, we also plan to extend the algorithm to be able to handle constraints, and for batch acquisition, i.e., where several solutions are to be selected in every iteration.


\section*{Acknowledgements}

The first author would like to acknowledge funding from ESTECO SpA and EPSRC through grant EP/L015374/1.

\printbibliography

@article{frazier2009knowledge,
  title={The knowledge-gradient policy for correlated normal beliefs},
  author={Frazier, Peter and Powell, Warren and Dayanik, Savas},
  journal={INFORMS journal on Computing},
  volume={21},
  number={4},
  pages={599--613},
  year={2009},
  publisher={INFORMS}
}

@article{pearce2022bayesian,
  title={Bayesian optimization allowing for common random numbers},
  author={Pearce, Michael Arthur Leopold and Poloczek, Matthias and Branke, Juergen},
  journal={Operations Research},
  year={2022},
  publisher={INFORMS}
}

@article{jones1998efficient,
  title={Efficient global optimization of expensive black-box functions},
  author={Jones, Donald R and Schonlau, Matthias and Welch, William J},
  journal={Journal of Global optimization},
  volume={13},
  number={4},
  pages={455--492},
  year={1998},
  publisher={Springer}
}

@inproceedings{rasmussen2003gaussian,
  title={Gaussian processes in machine learning},
  author={Rasmussen, Carl Edward},
  booktitle={Summer School on Machine Learning},
  pages={63--71},
  year={2003},
  organization={Springer}
}

@inproceedings{ES3_hernandez2014predictive,
  title={Predictive entropy search for efficient global optimization of black-box functions},
  author={Hern{\'a}ndez-Lobato, Jos{\'e} Miguel and Hoffman, Matthew W and Ghahramani, Zoubin},
  booktitle={Advances in neural information processing systems},
  pages={918--926},
  year={2014}
}

@inproceedings{wu2017bayesian,
  title={Bayesian optimization with gradients},
  author={Wu, Jian and Poloczek, Matthias and Wilson, Andrew G and Frazier, Peter},
  booktitle={Advances in Neural Information Processing Systems},
  pages={5267--5278},
  year={2017}
}

@inproceedings{poloczek2017multi,
  title={Multi-information source optimization},
  author={Poloczek, Matthias and Wang, Jialei and Frazier, Peter},
  booktitle={Advances in Neural Information Processing Systems},
  pages={4288--4298},
  year={2017}
}

@article{scott2011correlated,
  title={The correlated knowledge gradient for simulation optimization of continuous parameters using gaussian process regression},
  author={Scott, Warren and Frazier, Peter and Powell, Warren},
  journal={SIAM Journal on Optimization},
  volume={21},
  number={3},
  pages={996--1026},
  year={2011},
  publisher={SIAM}
}

@article{Thompson1933,
 ISSN = {00063444},
 author = {William R. Thompson},
 journal = {Biometrika},
 number = {3/4},
 pages = {285--294},
 publisher = {[Oxford University Press, Biometrika Trust]},
 title = {On the Likelihood that One Unknown Probability Exceeds Another in View of the Evidence of Two Samples},
 volume = {25},
 year = {1933}
}

@article{wu2017discretization,
  title={Discretization-free knowledge gradient methods for bayesian optimization},
  author={Wu, Jian and Frazier, Peter I},
  journal={arXiv preprint arXiv:1707.06541},
  year={2017}
}

@inproceedings{snoek2012practical,
  title={Practical bayesian optimization of machine learning algorithms},
  author={Snoek, Jasper and Larochelle, Hugo and Adams, Ryan P},
  booktitle={Advances in neural information processing systems},
  pages={2951--2959},
  year={2012}
}

@article{balandat2020botorch,
  title={BoTorch: A framework for efficient Monte-Carlo Bayesian optimization},
  author={Balandat, Maximilian and Karrer, Brian and Jiang, Daniel and Daulton, Samuel and Letham, Ben and Wilson, Andrew G and Bakshy, Eytan},
  journal={Advances in Neural Information Processing Systems},
  volume={33},
  year={2020}
}

@article{pearce2018continuous,
  title={Continuous multi-task Bayesian Optimisation with correlation},
  author={Pearce, Michael and Branke, Juergen},
  journal={European Journal of Operational Research},
  volume={270},
  number={3},
  pages={1074--1085},
  year={2018},
  publisher={Elsevier}
}

@inproceedings{wang2017max,
  title={Max-value entropy search for efficient Bayesian optimization},
  author={Wang, Zi and Jegelka, Stefanie},
  booktitle={Proceedings of the 34th International Conference on Machine Learning-Volume 70},
  pages={3627--3635},
  year={2017},
  organization={JMLR. org}
}

@article{hennig2012entropy,
  title={Entropy search for information-efficient global optimization},
  author={Hennig, Philipp and Schuler, Christian J},
  journal={Journal of Machine Learning Research},
  volume={13},
  number={Jun},
  pages={1809--1837},
  year={2012}
}

@article{kingma2014adam,
  title={Adam: A method for stochastic optimization},
  author={Kingma, Diederik P and Ba, Jimmy},
  journal={arXiv preprint arXiv:1412.6980},
  year={2014}
}

@article{Scott2011a,
	Author = {Scott, W. and Frazier, P. and Powell, W.},
	Issn = {1052-6234},
	Journal = {SIAM Journal on Optimization},
	Language = {en},
	Number = {3},
	Pages = {996--1026},
	Publisher = {Society for Industrial and Applied Mathematics},
	Title = {The Correlated Knowledge Gradient for Simulation Optimization of Continuous Parameters using {G}aussian Process Regression},
	Volume = {21},
	Year = {2011}}

@article{Frazier2008,
	Author = {Frazier, P. I. and Powell, W. B. and Dayanik, S.},
	Isbn = {10.1137/070693424},
	Journal = {SIAM Journal on Control and Optimization},
	Language = {en},
	Month = {sep},
	Number = {5},
	Pages = {2410--2439},
	Publisher = {Society for Industrial and Applied Mathematics},
	Title = {A Knowledge-Gradient Policy for Sequential Information Collection},
	Volume = {47},
	Year = {2008}}

@article{Frazier2009,
	Author = {Frazier, P. and Powell, W. and Dayanik, S.},
	Issn = {10919856},
	Journal = {INFORMS Journal on Computing},
	Number = {4},
	Pages = {599--613},
	Title = {The knowledge-gradient policy for correlated normal beliefs},
	Volume = {21},
	Year = {2009}}

@misc{pearce2020practical,
      title={Practical {B}ayesian Optimization of Objectives with Conditioning Variables}, 
      author={Michael Pearce and Janis Klaise and Matthew Groves},
      year={2020},
      eprint={2002.09996},
      archivePrefix={arXiv},
      primaryClass={stat.ML}
}

@inproceedings{ankenman2008stochastic,
  title={Stochastic kriging for simulation metamodeling},
  author={Ankenman, Bruce and Nelson, Barry L and Staum, Jeremy},
  booktitle={2008 Winter Simulation Conference},
  pages={362--370},
  year={2008},
  organization={IEEE}
}

@article{yamawaki2018multifunctional,
  title={Multifunctional structural design of graphene thermoelectrics by Bayesian optimization},
  author={Yamawaki, Masaki and Ohnishi, Masato and Ju, Shenghong and Shiomi, Junichiro},
  journal={Science advances},
  volume={4},
  number={6},
  pages={eaar4192},
  year={2018},
  publisher={American Association for the Advancement of Science}
}

@article{liu1989limited,
  title={On the limited memory BFGS method for large scale optimization},
  author={Liu, Dong C and Nocedal, Jorge},
  journal={Mathematical programming},
  volume={45},
  number={1},
  pages={503--528},
  year={1989},
  publisher={Springer}
}

@misc{shewchuk1994introduction,
  title={An introduction to the conjugate gradient method without the agonizing pain},
  author={Shewchuk, Jonathan Richard and others},
  year={1994},
  publisher={Carnegie-Mellon University. Department of Computer Science Pittsburgh}
}

\clearpage
\appendix

\section{One Step Look-Ahead Posterior Mean Derivation}\label{apndx:one_step_post_mean}
At iteration $n$ during optimization, let the training inputs be 
$X^n=\left(x^1,...,x^n\right)$ and
the training outputs $Y^n = (y^1,...,y^n)$. Given a prior mean and kernels functions, $\mu^0(x):X\to \R$ and $k^0(x,x'): X \times X \to \R$.
Finally let the new sample point be $x\n1$. Updating the mean
function with data from the $0^{th}$ step to $n^{th}$ step is given by
\begin{eqnarray}
    \mu^n(x) &=& \mu^0(x) + k^0(x, X^n) \underbrace{K^{-1}\left(Y^n - \mu^0(X^n)\right )}_{\text{define }\tilde Y^n} 
\end{eqnarray}
where $K=k^0(X^n, X^n)+\sigma_\epsilon^2I$. 

A simple change of indices from $0\to n$ and $n \to n+1$, yields the one-step updated posterior mean
\begin{equation} \label{eqn:one-mean}
    \mu\n1(x) = \mu^n(x) + \frac{k^n(x, x\n1)}{k^n(x\n1, x\n1)+\sigma_\epsilon^2}
    \left(y\n1 - \mu^n(x\n1)\right ).
\end{equation}
which contains the random $y\n1$ which has a predictive distribution 
\begin{equation}
    \P[y\n1|x\n1, X^n, Y^n] = N(\mu^n(x\n1), k^n(x\n1, x\n1)+\sigma_\epsilon^2).
\end{equation}
hence we may take factorise the one-step look head posterior mean expression as follows
\begin{eqnarray}
    \mu\n1(s,x) &=& \mu^n(s,x) + 
    k^n(x, x\n1)\frac{1}{\underbrace{\sqrt{ k^n(x\n1, x\n1)+\sigma_\epsilon^2}
    }_{\text{standard deviation of $y\n1$}}}
    \underbrace{\frac{\left(y\n1 - \mu^n(x\n1)\right )}
    {\sqrt{ k^n(x\n1, x\n1)+\sigma_\epsilon^2}}
    }_{\text{Z-score of $y\n1$}} \\
    &=& \mu^n(x) + \frac{
        k^n(x, x\n1)
    }{
        \sqrt{ k^n(x\n1, x\n1)+\sigma_\epsilon^2}
    }Z \label{eq:factorized_new_post_mean} \\
    &=& \mu^n(x) + \tilde\sigma(x, x\n1) Z
\end{eqnarray}
where the left factor is a deterministic and the right factor is
the (at time $n$) stochastic Z-score of the new $y\n1$ value. One may simply sample
$Z\sim N(0,1)$ values and compute Equation \ref{eq:factorized_new_post_mean} to
generate posterior mean functions.

\section{Discrete KG Algorithm}\label{apndx:discrete_KG_epigraph}
\newcommand{\mul}{\underline{\mu}}
\newcommand{\sigl}{\underline{\sigma}}
\newcommand{\Zt}{\underline{\tilde{Z}}}
\begin{algorithm}[!h] \caption{Knowledge Gradient by discretization. This algorithm takes as input a set of linear
functions parameterised by a vector of intercepts $\mul$ and a vector of
gradients $\sigl$. It then computes the intersections
of the piece-wise linear epigraph (ceiling) of the functions and 
the expectation of the output of the function given Gaussian input. 
Vector indices are assumed to start from 0.
\label{alg:KGdisc}}
\begin{algorithmic}
\Require $\mul$, $\sigl \in \R^{n_A}$
\State $O \gets \text{order}(\sigl)$                                        \Comment{get sorting indices of increasing $\sigl$}
\State $\mul \gets \mul[O]$, $\sigl \gets \sigl[O]$                         \Comment{arrange elements}
\State $I\gets[0,1]$                                                        \Comment{indices of elements in the epigraph}
\State $\Zt \gets [-\infty, \frac{\mu_0 - \mu_1}{\sigma_1 - \sigma_0}]$ \Comment{z-scores of intersections on the epigraph}
\For{$i=2$ \textbf{to} $n_z-1$}
	\State ($\star$) 
	\State $j\gets last(I)$ 
	\State $z\gets \frac{\mu_i - \mu_j}{\sigma_j - \sigma_i} $
	\If {$z<last(\Zt)$}
		\State Delete last element of $I$ and of $\Zt$
		\State Return to ($\star$)
	\EndIf
	\State Add $i$ to end of $I$ and $z$ to $\Zt$
\EndFor
\State $\Zt\gets [\Zt,\infty]$
\State $\underline{A} \gets \phi(\Zt[1:]) - \phi(\Zt[:-1])$ \Comment{assuming python indexing}
\State $\underline{B} \gets \Phi(\Zt[1:]) - \Phi(\Zt[:-1])$ 
\State $\KG \gets \underline{B}^T\mul[I] - \underline{A}^T\sigl[I] - \max \mul$ \Comment{compute expectation}
\State \Return KG
\end{algorithmic}
\end{algorithm}

\label{apndx:}

\end{document}